# Transfer learning for process design with reinforcement learning

Qinghe Gao[a], Haoyu Yang[a], Shachi M. Shanbhag[a], Artur M. Schweidtmann[a,*]

[a]*Process Intelligence Research Team, Department of Chemical Engineering, Delft University of Technology, Van der Maasweg 9, Delft 2629 HZ, The NETHERLANDS*
[*]*Corresponding author. Email: a.schweidtmann@tudelft.nl*

**Abstract**
Process design is a creative task that is currently performed manually by engineers. Artificial intelligence provides new potential to facilitate process design. Specifically, reinforcement learning (RL) has shown some success in automating process design by integrating data-driven models that learn to build process flowsheets with process simulation in an iterative design process. However, one major challenge in the learning process is that the RL agent demands numerous process simulations in rigorous process simulators, thereby requiring long simulation times and expensive computational power. Therefore, typically short-cut simulation methods are employed to accelerate the learning process. Short-cut methods can, however, lead to inaccurate results. We thus propose to utilize transfer learning for process design with RL in combination with rigorous simulation methods. Transfer learning is an established approach from machine learning that stores knowledge gained while solving one problem and reuses this information on a different target domain. We integrate transfer learning in our RL framework for process design and apply it to an illustrative case study comprising equilibrium reactions, azeotropic separation, and recycles, our method can design economically feasible flowsheets with stable interaction with DWSIM. Our results show that transfer learning enables RL to economically design feasible flowsheets with DWSIM, resulting in a flowsheet with an 8% higher revenue. And the learning time can be reduced by a factor of 2.

## 1. Introduction

The transition of chemical engineering to a sustainable and circular future requires new methods of process design (Fantke et al., 2021). Currently, methods for process design are mainly manual work with long development times and superstructure methods are also limited to predefined process configurations (Chen et al. 2017, Mitsos et al. 2018). Recently, reinforcement learning (RL), a branch of machine learning (ML), has shown promising results in process design (Midgley et al., 2020, Khan et al., 2020, Göttl et al., 2021, Stops et al., 2022, Kalmthout et al., 2022). One major challenge in RL for process design is the training process as it is trial-and-error based. Thereby, the learning process typically requires a large number of process simulations, which demands expensive computational power. Previous work (Khan et al., 2020, Göttl et al., 2021, Stops et al., 2022) mostly leverages short-cut process simulation methods to simulate the processes efficiently, which can lead to inaccurate results. Recent works employ rigorous process simulators such as COCO and Aspen Plus with rigorous models (Midgley et al., 2020, Kalmthout et al., 2022). Nevertheless, the problem of long simulation times hinders further developments (Kalmthout et al., 2022).

We propose to utilize transfer learning in process design with RL to facilitate the effectiveness and efficiency of the learning process. Transfer learning is a technique to improve learning performance by transferring knowledge from different but relevant domains to the target domain (Zhu et al., 2020). Adapting the concept of transfer learning for process design with RL, we first pre-train our recently proposed RL agent on a short-cut process simulation from our previous work (Stops et al., 2022). Then, we transfer the pre-trained agent to a rigorous process simulator DWSIM for further training. Finally, we illustrate the impact of transfer learning through one case study.

## 2. Methods

The RL problem can be formulated as a Markov decision process (MDP): $M = \{S, A, T, R\}$, which includes states $s \in S$, actions $a \in A$, transitions $T: S \times A$, and reward functions $R$. The agent aims to maximize the reward in the environment by literately taking action, evaluating the current reward, and updating the states. Specifically, in process synthesis tasks, states correspond to flowsheets. Actions are composed of two parts: Discrete and continuous actions. The discrete actions include the selections of a unit operation and its location in the flowsheet. The continuous actions define the design and operation variables of the corresponding unit operation. After the agent has performed actions, the states are updated. Based on the current state, a reward is calculated by the environment, e.g., the process simulation software, and returned to the agent as feedback on its actions. For process design, this reward is typically a design goal such as the process revenue. By repeating the process of performing actions and receiving rewards multiple times, the agent is trained to perform actions that result in a higher reward, corresponding to flowsheets with higher revenue.

*2.1. Agent and environment*

We adapt the RL framework from our previous work (Stops et al., 2022), where the states are presented as directed flowsheet graphs. Within the directed flowsheet graphs, nodes represent the unit operations and edges correspond to the process streams. Each node and edge is assigned a feature vector, respectively, storing information about the unit operation, e.g., type or size, and stream, e.g., thermodynamic data or flow rate of the stream. Furthermore, the agent architecture consists of three major parts: Graph encoder, actor networks, and critic networks. The graph encoder takes current flowsheet graphs as input and utilizes graph convolution in GNNs to learn information about the flowsheet graphs, in form of a vector representation, also referred to as flowsheet fingerprint. Actor networks are responsible for taking actions during the training process. There are three action levels: Selecting an open stream, selecting a unit operation, and selecting a corresponding design variable. Moreover, taking the flowsheet fingerprint as input, critic networks are used to estimate the reward of actions, and then actor networks will learn to take actions with the highest estimated reward. Specifically, the reward is calculated by using the DWSIM process simulator (Medeiros et al., 2018). The RL framework is implemented in Python including an interface for the agent to actively interact with DWSIM.

*2.2. Transfer learning*

We extend our RL framework by transfer learning. Specifically, we add a pre-training phase to the training of the agent. In the pre-training phase, we use a short-cut process simulation environment (Stops et al., 2022) and train the agent over 10000 episodes. Then, we transfer the pre-trained agent to a fine-tuning phase in which the agent is trained against a rigorous process simulator DWSIM for further 15000 training episodes. For the comparison, we directly train another agent with DWSIM environment over 15000 episodes. Note that in each episode, the agent generates a complete flowsheet. Both pre-training and fine-tuning processes are adapted from Proximal Policy Optimization (PPO) by OpenAI (Schulman et al., 2017). Then, the agent architecture is updated by gradient descent for the total loss function derived from summing up losses of actor networks, loss of critic networks, and corresponding entropies.

The agents are trained on a Windows server with a 3.5 GHz 24 cores Intel(R) Xeon(R) W-2265 CPU, NVDIA GeForce RTX 3090 GPU and 64 GB memory.

## 3. Illustrative case study

The production of methyl acetate (MeOAC) is chosen as an illustrative case study.

*3.1. Process simulation*

The short-cut process methods for pre-training are illustrated in our previous work (Stops et al., 2022). Here, the process simulation with DWSIM is introduced. In this case study, the agent can choose reactors, distillation columns, and heat exchangers as unit operations. Besides, the agent can also decide to add recycles or claim open streams as products. The types of unit operations and corresponding design variables are defined as follows.

**Reactor** is deployed to convert reactants to the desired product (MeOAc). The reactor is modeled as a plug flow reactor (PFR) where the following reversible reaction takes place:

$$HOAc + MeOH \rightleftharpoons MeOAc + H_2O \quad (1)$$

For operational simplicity, the reactor is simulated isothermally, in which the temperature is kept constant regarding the inlet stream temperature. Besides, catalyst loading is not considered. The reaction kinetics are obtained from (Xu et al, 1996), with the equilibrium being related to temperature. The reactor cross-sectional area is determined by the relation: N/10, where N is the inlet molar flow (Stops et al., 2022). The design variable is the reactor length, which will be determined by the agent in the third-level continuous decision process. The range is from 3 to 10 m.

**Distillation column** is applied to separate MeOAc from the quaternary system. Rigorous distillation columns are used instead of shortcut columns in the previous work (Stops et al., 2022) to account for more realistic factors such as intermolecular interactions. The rigorous column models provide multiple possible choices of design parameters, from which the distillate to feed ratio (D/F) is selected as the third-level decision. Other adjustable parameters such as the number of stages and reflux ratio are set as fixed values (35 and 1.5, respectively). The D/F ratio can range from 0.4 to 0.6.

**Heat exchanger** is a DWSIM heater model. In the proposed framework, the heat exchanger is simulated based on the outlet temperature which is determined by the third-level decision. A temperature range from 278.15 K to 330.05 K is applied, where the upper limit refers to the lowest boiling point of the components which is MeOAc.

**Recycle** action consists of additional units including a splitter and a mixer. Firstly, the process stream is split into a recycle stream and a purge. Secondly, the recycled stream is merged with the selected feed with a mixer. Thereby, the split ratio is the third-level decision of the agent, which lies in the range of 0.1 to 0.9 for pragmatic consideration.

*3.2. Reward*

The reward determines the economic viability of generated flowsheets and teaches the agent to take feasible actions. First, a reward of 0 € is given when the incomplete flowsheets can converge after every single action, because the economic value is difficult to assess for an incomplete flowsheet. Second, whenever the agent fails the simulation by taking infeasible actions, the episode will be terminated immediately and a negative reward -10M€ is given. Finally, when a flowsheet is completed, we calculate the reward according to Equation 2:

$$r = \sum P_{product} - \sum C_{feed} - \sum (C_{operation} + 0.15 \cdot C_{invest})_{units} \quad (2)$$

where $P_{product}$ is the revenue of the sold product (Seider et al., 2008), $C_{feed}$ is the costs of feeds, $C_{opeartion}$ is the operation costs (Smith, 2016) and $C_{invest}$ is the total capital investment which is multiplied by factor 0.15 (Seider et al., 2008). In the case of negative rewards, a reduction factor 10 is applied to encourage the exploration of the agent.

## 4. Results and discussion

Figure 1 shows the learning curves for the agents with and without transfer learning. The scores represent the moving average rewards, i.e., the economic viability of the flowsheets, over 100 episodes. The training for the agent without transfer learning took 72 hours over 10000 episodes. For the agent with transfer learning, the pre-training took 2 hours over 10000 episodes and the further training took 72 hours over 10000 episodes. During the first 3500 episodes, the agent without transfer learning generates predominantly infeasible lengthy flowsheets, resulting in the learning curves rising slowly. In fact, due to the complexity of the design space, the agent has difficulty in learning from the previous failed flowsheets, leading to negative scores. After 3500 episodes, the agent mainly produces flowsheets with positive scores, which indicates that the designed processes are economically viable. Besides, within the training episodes, the learning curve slowly converges and reaches maximally to about 42. In comparison, the agent with transfer learning shows a quicker learning process. At the beginning of the learning process, the agent is able to mostly produces positive scores, and then the learning curve rises steeply. This demonstrates that the agents successfully leverage the pre-trained information from a short-cut process environment to make favorable decisions even in the early training stages. After about 4500 episodes, the score begins to fluctuate between 30 and 40 and maximumly reaches 46 which is higher than the agent without transfer learning.

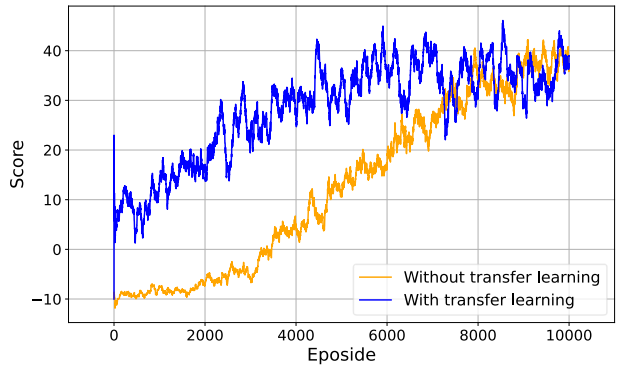

Figure 1: Learning curves of the agent with and without transfer learning. The blue line depicts the learning curve with transfer learning and the orange line indicates the learning curves without transfer learning. The scores are moving average rewards over 100 episodes and each learning curve runs over 10000 episodes.

Figure 2 displays the best flowsheet generated by the transfer learning agent and the continuous design variables are shown in Table 1. First, the feed (F1) is fed directly into three consecutive reactors (R1, R2, and R3) where the MeOAC and $H_2O$ are produced by the esterification of HOAC and MeOH. Then, the resulting quaternary mixture is split up in a column (C1). The mixture is distilled from the top part of the column and sent to one heat exchanger (Hex1) to get the product (P1), containing enriched MeOAC and residues of HOAC and $H_2O$. The bottom product of the first column is further split up into the second column (C2) to produce pure $H_2O$ in the distillate (P2) and a mixture of MeOH and $H_2O$ in the third product stream (P3). And 90 % of the bottom product is recycled and mixed back into the feed.

While the resulting flowsheet has a positive reward, it is still far from a realistic engineering solution and future research is required. In particular, there are three major issues observed in the optimal flowsheet solution: (1) In industrial applications, MeOAC is primarily produced in reactive distillation (Huss et al., 2003). As our agent does not include reactive distillation as a unit operation, this cannot be identified. In future work, intensified unit operations can be added to our framework or the agent could operate on a phenomena level rather than a unit operation level. (2) The best flowsheet generated by the agent in this work contains three consecutive PFRs. The reason is that the length of a single PFR is limited to 10 m, which is not sufficient to finish the reaction. Therefore, the agent learns to choose multiple PFRs to fulfill the reaction and maximize the product. (3) One unnecessary heat exchanger is added after the distillation column C1 and before the product P1. We believe this is due to the small operations cost of the heat exchangers and the minor impact on the overall reward. Future research should further investigate possible mitigation strategies such as longer training or further hyperparameter tuning.

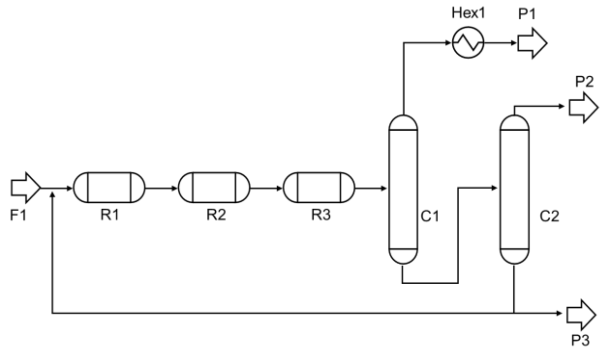

Figure 2: Best flowsheet generated by the transfer learning agent. First, MeOAC and H2O are produced from the feed (F1) in three consecutive reactors (R1, R2, and R3). Then the mixture is separated in the first column (C1). The first product (P1) is enriched with MeOAc but also contains residues of HOAC and H2O after one heat exchanger (Hex1). Then the bottom mixture of MeOH and H2O is further separated in the second column (C2). Pure H2O is in the distillate (P2) and the third product (P3) is the mixture of MeOH and H2O. And 90% of P3 is recycled and mixed with the feed stream.

Table 1. Design variables selected by the agent with transfer learning in the best flowsheet.

| Unit operation | Design variable | Unit | Best run |
|---|---|---|---|
| Reactor 1 (R1) | Length | m | 9.42 |
| Reactor 2 (R2) | Length | m | 9.25 |
| Reactor 3 (R3) | Length | m | 8.38 |
| Column 1 (C1) | Distillate to feed ratio | - | 0.58 |
| Column 2 (C2) | Distillate to feed ratio | - | 0.4 |
| Heat exchanger 1 (Hex1) | Outlet temperature | K | 315 |
| Recycle | Recycled ratio | - | 0.9 |

## 5. Conclusion

We propose to deploy the transfer learning for process design in RL to accelerate the learning process of the agent. The GNNs-based agent is first pre-trained with a short-cut simulation environment and then transferred to the rigorous process simulator environment for further training. In the illustrative case study, the agent is able to design economically feasible flowsheets in the process simulator DWSIM environment. Furthermore, the learning curves demonstrate that transfer learning indeed improves the efficiency of the learning process significantly and thus can be used to reduce the overall training time significantly. This work thus demonstrates that transfer learning can accelerate the learning process of graph-based RL with rigorous process simulator environments.